\documentclass[10pt,letterpaper]{article}
\usepackage{authblk}
\usepackage{cogsci}
\usepackage{pslatex}
\usepackage{apacite}
\usepackage{linguex}
\usepackage{graphicx}
\usepackage{xcolor}
\usepackage{amsmath}
\usepackage{tikz}
\usepackage{tikz-qtree}
\usepackage{float}

\title{What Syntactic Structures block Dependencies in RNN Language Models?}
 
\author[1]{\textbf{Ethan Wilcox}}
\author[2]{\textbf{Roger Levy}}
\author[3]{\textbf{Richard Futrell}}

\affil[1]{Department of Linguistics, Harvard University, \tt{wilcoxeg@g.harvard.edu}}
\affil[2]{Department of Brain and Cognitive Sciences, MIT, \tt{rplevy@mit.edu}}
\affil[3]{Department of Language Science, UC Irvine, \tt{rfutrell@uci.edu}}

\begin{document}
\setlength{\Exlabelwidth}{0.7em}
\setlength{\Exlabelsep}{0.9em}
\setlength{\SubExleftmargin}{1.3em}
\setlength{\Extopsep}{1pt}

\maketitle

\begin{abstract}

Recurrent Neural Networks (RNNs) trained on a language modeling task have been shown to acquire a number of non-local grammatical dependencies with some success \cite{linzen2016assessing}. Here, we provide new evidence that RNN language models are sensitive to hierarchical syntactic structure by investigating the \textbf{filler--gap dependency} and constraints on it, known as \textbf{syntactic islands}. Previous work is inconclusive about whether RNNs learn to attenuate their expectations for gaps in island constructions in particular or in \emph{any} sufficiently complex syntactic environment. This paper gives new evidence for the former by providing control studies that have been lacking so far. We demonstrate that two state-of-the-art RNN models are are able to maintain the filler--gap dependency through unbounded sentential embeddings and are also sensitive to the hierarchical relationship between the filler and the gap.  Next, we demonstrate that the models are able to maintain \textbf{possessive pronoun gender expectations} through island constructions---this control case rules out the possibility that island constructions block all information flow in these networks. We also evaluate three untested islands constraints: coordination islands, left branch islands, and sentential subject islands. Models are able to learn left branch islands and learn coordination islands gradiently, but fail to learn sentential subject islands. Through these controls and new tests, we provide evidence that model behavior is due to finer-grained expectations than gross syntactic complexity, but also that the models are conspicuously un-humanlike in some of their performance characteristics.


\end{abstract}

\section{Introduction}

Recurrent Neural Networks (RNNs) with Long Short-Term Memory architecture (LSTMs) have achieved state-of-the-art scores at a number of natural language processing tasks, including language modeling and parsing \cite{hochreiter1997long,jozefowicz2016exploring}. In addition, they have begun to be used as a plausible sub-symbolic model for a variety of cognitive functions, including visual perception and language processing and comprehension \cite{elman1990finding}. However, the distributed representations learned by RNNs and neural networks in general are notoriously opaque, posing a challenge for their interpretability as models of human sentence processing and for their controllability as NLP systems.

One recent line of work aims to uncover what these `black boxes' learn about language by treating them like human psycholinguistic subjects. In this \textbf{psycholinguistic paradigm} RNNs trained on the language modeling task are fed hand-crafted sentences, designed to expose their underlying syntactic knowledge \cite{linzen2016assessing,mccoy2018revisiting}. Much of this work has investigated what RNNs trained on a language modeling objective are capable of learning about natural syntactic dependencies. For the purposes of this investigation, we define \textbf{dependency} as any systematic co-variation between two words. For example, in one experiment networks were tested as to whether they had learned the number agreement dependency between a subject and a verb. They were fed with the prefix \textit{The key to the cabinet...} and correctly gave a higher probability to the grammatical \textit{is} over the ungrammatical \textit{are}. Networks were shown to successfully complete this task for a number of languages, as well as for sentences whose content words were replaced with random alternatives of the same syntactic category rendering them syntactically licit but semantically implausible \cite{gulordava2018colorless}.

But learning that covariance exists between certain words or word forms, without reference to their relative positions, is not enough to say that the RNN models have fully learned a dependency. Natural language dependencies consist of co-variation between two elements in \textit{certain syntactic positions}. Agents must both attend to the structural relationship between the two elements bound by the dependency and filter out intervening material in syntactically irrelevant positions. The subject--verb number agreement task above provides compelling evidence that RNNs are capable of the latter: they were able to maintain correct predictions despite a number of \textit{distractors} that mismatched the subject in number, such as \textit{cabinet} in the example provided \cite{marvin2018targeted}.

Evidence suggesting that RNN language models are also sensitive to the structural relationship between the two bound elements has emerged from the study of \textbf{filler--gap dependencies} \cite{wilcox2018rnn,chowdhury2018rnn}. The filler--gap dependency is the dependency between a \textit{filler}---such as \textit{who} or \textit{what}---and and a gap, which is an empty syntactic position. Crucially, filler--gap dependencies are subject to a number of constraints, known as \textit{island constraints}, which are a set of structural positions that prevent the filler and the gap from entering into a dependency with each other \cite{ross1967constraints}. \ref{ex:wh-island-intro-island} gives one example island, in which the dependency is blocked by a wh-complementizer.

\ex. \label{ex:wh-island-intro}
\a. I know what the guide said that the lion devoured \_\_ yesterday.  \textsc{\small{no violation}}
\b. *I know what the guide said whether the lion devoured \_\_ yesterday. \textsc{\small{wh-island island violation}} \label{ex:wh-island-intro-island}

While it has been shown that both simple Elman RNNs and more contemporary LSTMs are able to represent the basic covariance between fillers and gaps, as well as other non-structural aspects of dependency, it is still uncertain whether the models are sensitive to island constraints \cite{elman1991distributed}. Previous work has demonstrated that two state-of-the-art models are sensitive to three of the most-studied island constraints (wh-islands, complex NP islands and adjunct islands) but insensitive to a fourth (subject islands) \cite{wilcox2018rnn}. Others have concluded that the models are merely sensitive to syntactic complexity plus order.  \citeA{chowdhury2018rnn} compared sentence-level perplexity scores obtained by RNN LMs for wh-questions that violate island constraints, and yes-no questions and statements that violate no grammatical rules but contain the same syntactic structures. While the models obtained better perplexity scores on the statements compared to the island-violation questions, they performed similarly on the island-violations and non-violating yes/no questions. These results may indicate that RNNs are not learning to attenuate their expectations for gaps in island constructions in particular, but in \emph{any} sufficiently complex syntactic environment.

This paper adjudicates between these two accounts of model behavior by providing control studies that have been lacking so far. In the first section, we demonstrate that two state-of-the-art LSTM models are sensitive to some forms of syntactic complexity, but not to others. Models are able to maintain the filler--gap dependency through \textbf{unbounded sentential embeddings} and yet are sensitive to the \textbf{hierarchical relationship} between the filler and the gap, suggesting that only specific types of syntactic complexity block gap expectations. In the second section, we turn to \textbf{possessive pronoun gender dependencies}, demonstrating that the models are able to maintain general expectations through island constructions---it is not the case that island constructions block all information flow in these networks. In this section we also evaluate three untested islands constraints: \textbf{coordination islands}, \textbf{left branch islands}, and \textbf{sentential subject islands}. Models are able to learn left branch islands and coordination islands gradiently, but fail to learn sentential subject islands. Through these controls and new tests, we provide evidence that model behavior is due to finer-grained expectations than gross syntactic complexity, but also that the models are conspicuously un-humanlike in some of their performance characteristics.

\vspace{-0.2cm}
\section{Methods}
\vspace{-0.2cm}

\subsection{Language Models}

We assess two state-of-the-art pre-existing LSTM models trained on English text for a language modeling objective. The first model, which we refer to as the \textbf{Google Model}, was trained on the One Billion Word Benchmark and has two hidden layers with 8196 units each. It uses the output of a character-level convolutional neural network (CNN) as input to the LSTM (and was originally presented as the \textit{BIG LSTM+CNN Inputs}) \cite{jozefowicz2016exploring}. The second model, which we refer to as the \textit{Gulordava Model} was selected for its previous success at learning the subject-verb number agreement task. It was trained on 90 Million tokens of English Wikipedia, and has two hidden layers of 650 units each \cite{gulordava2018colorless}.

\subsection{Dependent Measure: Surprisal}

In this work we take a grammatical dependency to be the co-variance between an upstream \textit{licensor} and a downstream \textit{licensee}. We assess the model's knowledge of the dependency by measuring the effect that the licensor has on the \textbf{surprisal} of the licensee, or on material immediately following the licensee when it is a gap. Surprisal, or negative log-conditional probability , $S(x_i)$ of a sentence's $i^{th}$ word $x_i$, tells us how strongly $x_i$  is expected under the language model's probability distribution. For sentences out of context, the surprisal is: $S(x_i) = -\log p(x_i|x_1 \dots x_{i-1})$. Surprisal is known to correlate directly with processing difficulty in humans \cite{smith2013effect, hale2001probabilistic, levy2008expectation}. In this work, we expect that grammatical licensors set up expectations for licensee, reducing its surprisal compared to minimal pairs in which the licensor is absent. We derive the word surprisal from the LSTM langauge model by directly computing the negative log of the predicted conditional probability $p(x_i|x_1 \dots x_{i-1})$ from the softmax layer. 

\subsection{Experimental Design: Wh-Licensing Interaction}

The filler--gap dependency is biconditional: Fillers set up expectations for gaps and gaps require fillers to be licensed. To measure this bi-directionality we employ the 2x2 interaction design proposed in \citeauthor{wilcox2018rnn}. There, the authors measure the \textbf{wh-licensing interaction}, which they compute from four sentence variants, given in \ref{ex:filler-gap-dep}, that contain the four possible combinations of fillers and gaps for a specific syntactic position. Note that the underscores are for presentational purposes only, and were not included in test items. Subsequent examples will be given via the \ref{ex:nofiller-gap} example, but all four variants were created in order to compute the licensing interaction.

\ex. \label{ex:filler-gap-dep} 
\small
\a. I know that you insulted your aunt yesterday. \textsc{\small[-Filler -Gap]} \label{ex:nofiller-nogap}
\b. *I know who you insulted your aunt yesterday. \textsc{\small[+Filler -Gap]} \label{ex:filler-nogap}
\c. *I know that you insulted \_\_ yesterday. \textsc{\small[-Filler +Gap]}\label{ex:filler-gap}
\d. I know who you insulted \_\_ yesterday. \textsc{\small[+Filler +Gap]}\label{ex:nofiller-gap}

If the filler sets up an expectation for a gap, then the filled syntactic position where a gap would typically occur should be more surprising in contexts that contain an upstream filler. That is $S(b)-S(a)$ should be a large positive number. If the gap requires a filler to be licensed, then the transition from the embedded verb to the S-modifying PP `yesterday' that skips over the otherwise-required grammatical object should be more surprising in contexts without an upstream filler. That is, $S(d)-S(c)$ should also be a large negative number. We can assess how well the model has learned both expectations by measuring the difference of differences: $[S(b)-S(a)]-[S(d)-S(c)]$. This is the wh-licensing interaction. If the models are learning the filler--gap dependency, we expect this to be a large positive number, with typical models showing about 4 bits of licensing interaction in simple object extracted clauses such as \ref{ex:filler-gap-dep}. Although we might expect the strongest difference in surprisal between \ref{ex:nofiller-nogap} and \ref{ex:filler-nogap} to be on the filled-gap position, \textit{your aunt}, this material is elided in two of the conditions. Therefore, in order to keep the measurement site the same across all four conditions, we measure wh-licensing interaction in the post-gap prepositional phrase (`yesterday' in \ref{ex:filler-gap-dep}).

In previous work using this methodology, RNN knowledge of island constraints was assessed by comparing the licensing interaction in island configurations to that in non-island minimal pairs. Strong evidence for an island constraint would be if the wh-licensing interaction dips to zero for a gap in island position, indicating that the model has decoupled expectations for fillers from gaps in this position. In practice we look for a significant decrease in wh-licensing interaction as indication that the models have learned to attenuate their expectations for gaps within islands. We derive the statistical significance of the interaction from a mixed-effects linear regression model, using some-coded conditions \cite{baayen2008mixed}. We include random intercepts by item but omit random slopes as we do not have repeated observations within items and conditions \cite{barr2013random}. In our figures, error bars represent 95\% confidence intervals of the contrasts between conditions, computed by subtracting out the by-item means before calculating the intervals as advocated in \cite{masson2003using}. \footnote{Our studies were preregistered on \url{aspredicted.org}: To see the preregistrations go to \url{aspredicted.org/blind.php?=$X$ where $X \in \{\texttt{sz8f5d}, \texttt{2r2eu7}, \texttt{zt73qt}, \texttt{es8rx7}, \texttt{f9pk9f}, \texttt{se6i2e}\} $.}}

\section{Syntactic Complexity}

\begin{figure}
    \centering
\begin{tikzpicture}
\tikzset{level distance=18pt}
\tikzset{sibling distance=2pt}
\Tree[.$\alpha$ 
    [.$\beta$ 
        [.\color{red}{$\gamma$} ]
        [.\color{blue}{$\delta$} 
            [.\color{blue}{$\epsilon$} ]
            [.\color{blue}{$\zeta$} ]
        ]
    ]
    [.$\eta$ ]
]
\end{tikzpicture}
\caption{C-Command in a binary-branching tree structure. $\gamma$ $c$-commands all the nodes in blue, but does not $c$-command the black nodes.}
    \label{fig:c-command}
\end{figure}
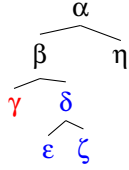

\subsection{Unboundedness} \label{sec:unboundedness}

\begin{figure*}
\begin{minipage}{0.32\textwidth}
\includegraphics[width=\textwidth]{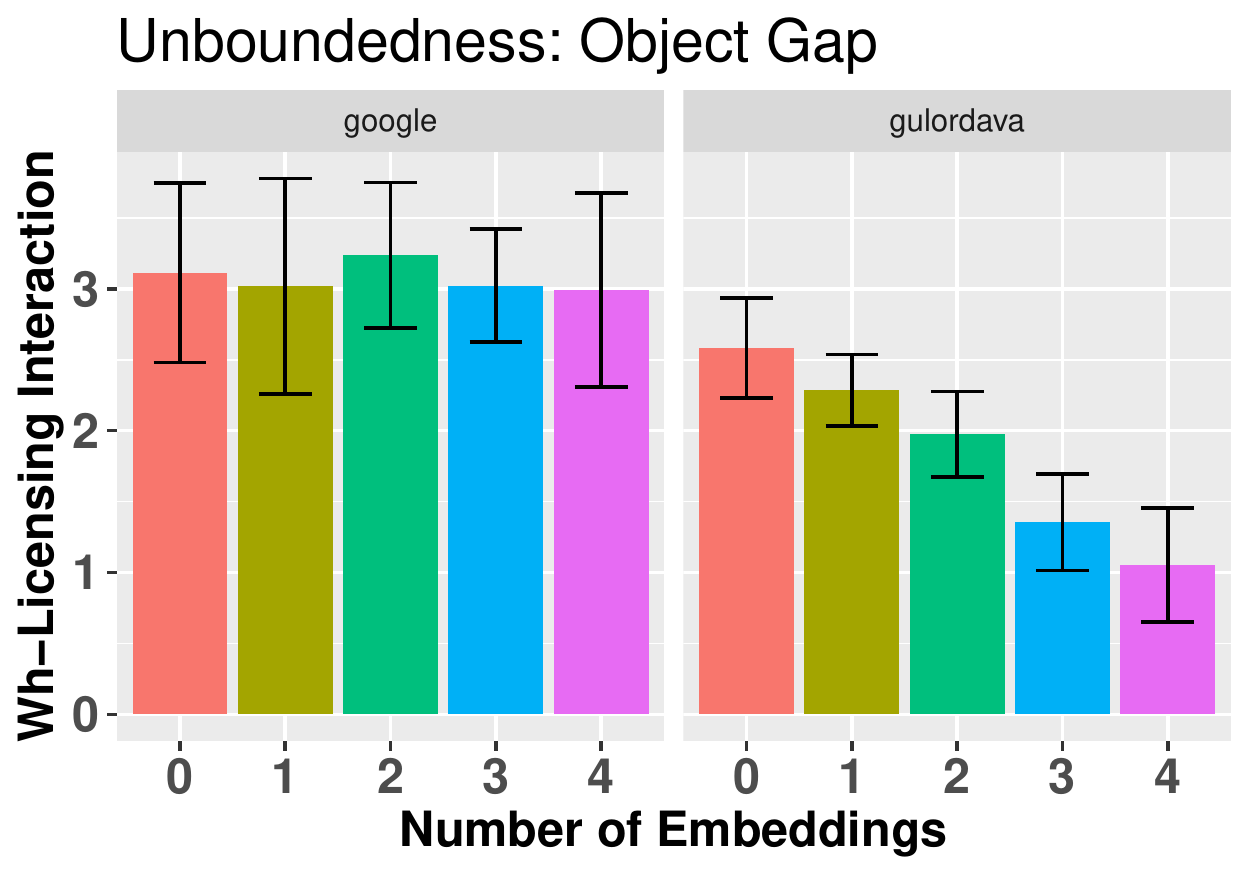}
\end{minipage}
\begin{minipage}{0.32\textwidth}
\includegraphics[width=\textwidth]{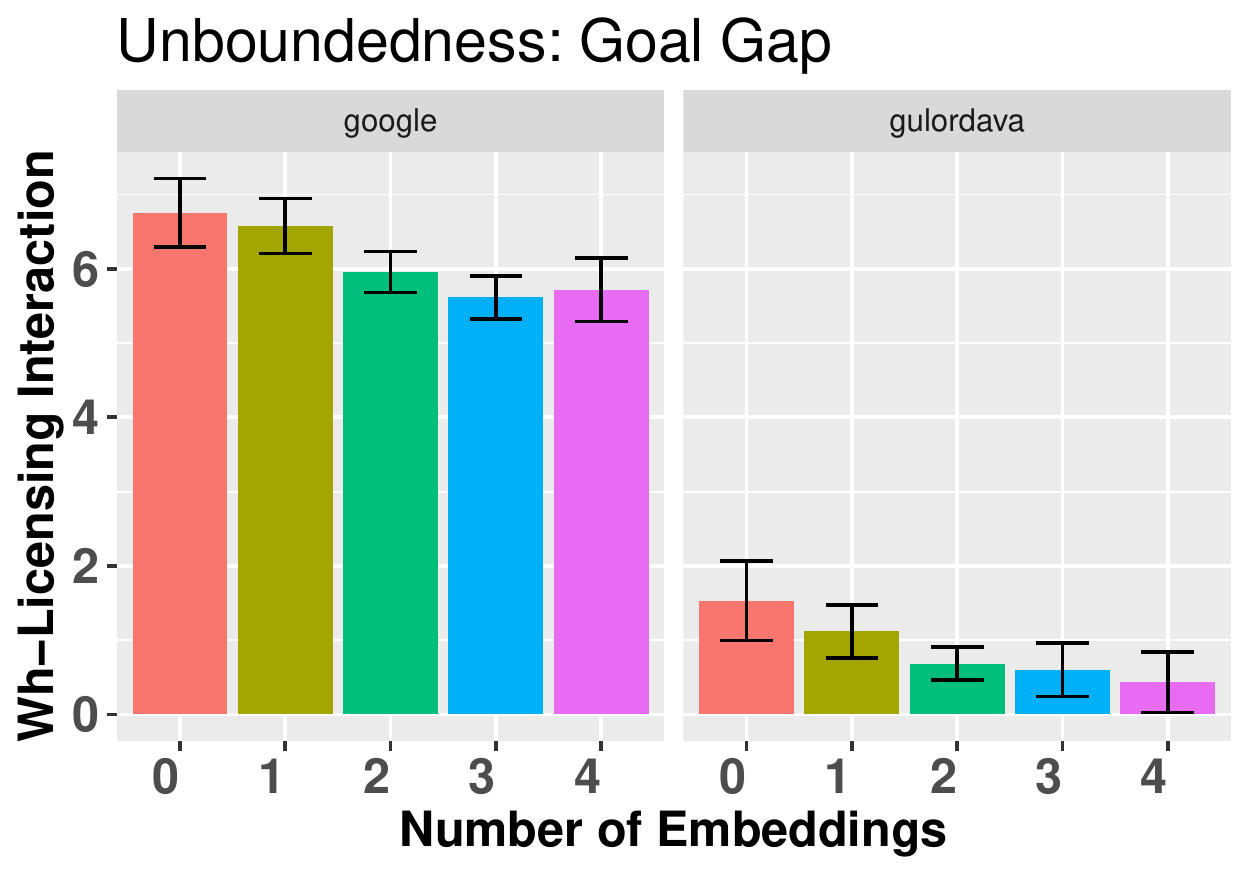}
\end{minipage}
\begin{minipage}{0.32\textwidth}
\includegraphics[width=\textwidth]{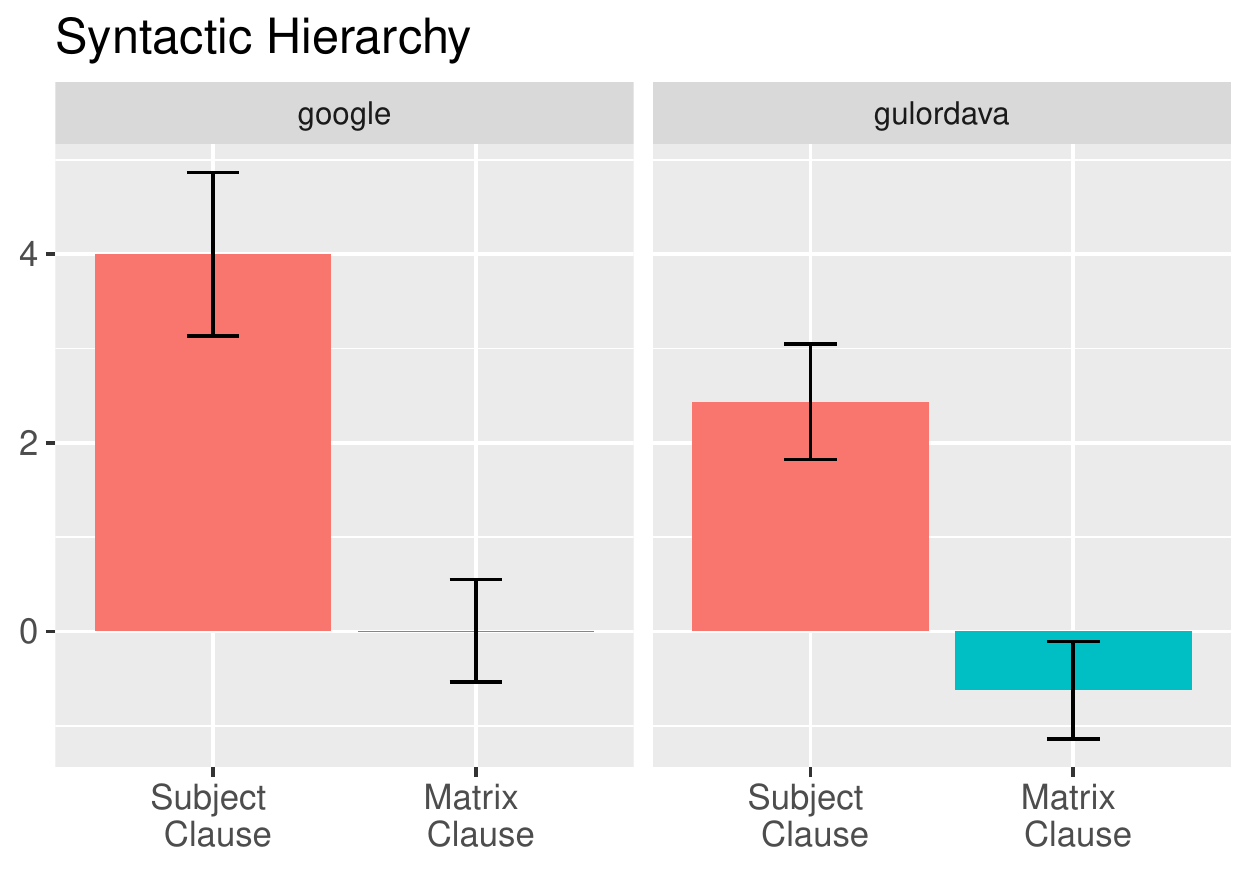}
\end{minipage}
\vspace{-0.3cm}
\caption{Effect of sentential embedding and syntactic hierarchy on wh-licensing interaction.}
\label{fig:unbound-hierarchy}
\vspace{-0.5cm}
\end{figure*}

The filler--gap dependency can span through a potentially unbounded number of sentential embeddings. To test whether models' expectations were attenuated with greater embedding depth, we created 23 items in five experimental conditions with between 0 and 4 layers of embedding and gaps in either object or indirect object (goal) position, following the examples in \ref{ex:unboundedness}, and measured the licensing interaction in the post-gap material. (In this and subsequent examples, the material in which the interaction is measured will be highlighted in bold.)

\ex. \label{ex:unboundedness}
\a. I know who you insulted \_\_ \textbf{at the party}. \textsc{\small[object gap, 0 layers]}
\b. I know who the gardener reported the butler said the hostess believed her aunt suspected you insulted \_\_ \textbf{at the party}. \textsc{\small[object gap, 4 layers]}
\c. I know who you delivered a challenge to \_\_ \textbf{at the party}. \textsc{\small[goal gap, 0 layers]}
\d. I know who the gardener reported the butler said the hostess believed her aunt suspected you delivered a challenge to \_\_ \textbf{at the party}. \textsc{\small[goal gap, 4 layers]}

The results for this experiment can be seen in figure \ref{fig:unbound-hierarchy}, with the object gap results on the top and goal gap results on the bottom. First, we find a significant interaction between fillers and gaps resulting in supperaditive reduction of surprisal ($p<0.001$ for all conditions) indicating that both models have learned the filler--gap dependency. Starting with the object gap conditions: For the google model, we find no effect of embedding depth on the wh-licensing interaction ($p>0.85$ in all cases); for the gulordava model, we find a significant decrease in wh-licensing interaction only between the \textit{no embedding} conditions and conditions with 3 or 4 additional layers of embedding ($p<0.001$ in both). When the gap occurs in the goal position, for the google model, we find no significant effect of embedding depth of the wh-licensing interaction. For the gulordava model, we find a generally smaller wh-licensing interaciton, as well as a significant effect of embedding between the \textit{no embedding} condition and conditions with two or more additional embedding layers ($p<0.05, p<0.05, p<0.01$ for 2 ,3 and 4 layers). We take these results to indicate that the google model has learned the unboundedness of the filler--gap dependency whereas the gulordava model has learned only relative unboundedness and shows behavior that reflects human performance more than human competence. However, these results indicate that both models can, in principle, thread their expectations for gaps through complex syntactic structures, if we take the number of syntactic nodes as a proxy measure for syntactic complexity.

\subsection{Syntactic Hierarchy}

Although the filler--gap dependency is unbounded, it is subject to a number of hierarchical constraints, the most basic of which is that the filler must be ``above" the gap, structurally. Here, we take this to mean that the filler must $c$-command the gap, although the precise relationship is more complex \cite{pollard1994headdriven}. Structurally-speaking node $\gamma$ $c$-commands node $\delta$ if neither node directly dominates the other and every node $X$ that dominates $\gamma$ also dominates $\delta$. Figure \ref{fig:c-command} demonstrates this relationship, with the noes $c$-commanded by $\gamma$ highlighted in blue.

To assess whether the models had learned this constraint on the structural relationship we created 24 variants following the examples in \ref{ex:hierarchy} and measured the wh-licensing interaction in the post-gap PP. If the model has learned the structural constraints on the filler--gap dependency, an undischarged filler in the matrix clause should not make a gap in subsequent parts of the sentence more or less likely, leading to near-zero licensing interaction in the \textit{Matrix Clause} condition.

\ex.  \label{ex:hierarchy}
\a. The fact that the mayor knows who the criminal shot \_\_ \textbf{shocked the jury} during the trial. \textsc{\small[Subject]} 
\b. *The fact that the mayor knows who the criminal shot the teller shocked \_\_ \textbf{during the trial}. \textsc{\small[Matrix]} 

The results from this experiment can be seen in Figure \ref{fig:unbound-hierarchy}, on the far right panel. We find strong licensing interaction for the grammatical \textit{Subject Clause} conditions (in red), but a striking reduction in licensing interaction for the \textit{Matrix Clause} conditions (in blue), which is significant for both models ($p<0.001$). As the results in \ref{sec:unboundedness} and \citeauthor{wilcox2018rnn} have shown that RNN models are insensitive to linear distance between the filler and the gap, we take these results suggest that it is the relevant structural properties which block the models' expectations for gaps inside the matrix clause.

\section{Island Effects: Gender Expectation vs. Filler--Gap Dependency}

Island constraints are specific syntactic configurations that block the filler--gap dependency. One way to show that the RNN models are learning island conditions as constraints on the filler--gap dependency is to demonstrate that they are capable of threading other expectations into island configurations. To do this, we used \textbf{pronoun gender expectation} between a gendered noun, such as `actress' or `husband', and a possessive pronoun such as `his' or `her.'. Nouns that carry overt gender marking or culturally-imbued gender bias set up expectations that subsequent pronominals match them in gender. Previous work has shown that humans thread expectations set up by \textit{cataphoric pronouns} into syntactic islands \cite{yoshida2014origin}. Cataphoric pronouns are pronouns that precede the nominal element to which they refer, as in \ref{ex:cataphor}.

\ex. \label{ex:cataphor}
\textbf{Her} manager revealed that the studio notified \textbf{Judy Dench} about the new film.

Because cataphoric pronouns are relatively less frequent than anaphoric pronouns, which follow the nominal to which they refer, we use sentences such as those in \ref{ex:gender-dep} to assess whether RNN LMs can thread expectations into island environments. We measure the strength of the gender expectation by calculating the difference in surprisal between the matching condition and the mismatching condition, or S(\ref{ex:gend-mismatch})-S(\ref{ex:gend-match}). If the models attenuate their expectation for gender agreement in island positions, then we expect an interaction between {\textsc{mismatch}} and {\textsc{island}} resulting in supperaditivally lower surprisal. 

\ex. \label{ex:gender-dep}
\a. The actress said that they insulted \textbf{her} friends. \textsc{\small[match, control]} \label{ex:gend-match}
\b. \# The actress said that they insulted \textbf{his} friends. \textsc{\small[mismatch, control]} \label{ex:gend-mismatch}
\c. The actress said whether they insulted \textbf{her} friends. \textsc{\small[match, island]}
\d. \# The actress said whether they insulted \textbf{his} friends. \textsc{\small[mismatch, island]}

\begin{figure*}
\centering
\includegraphics[width=0.16\textwidth]{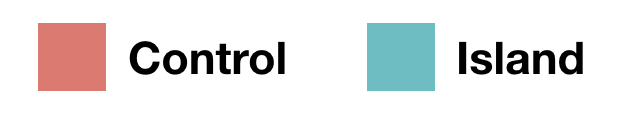}

\begin{minipage}{0.16\textwidth}
\includegraphics[width=\textwidth]{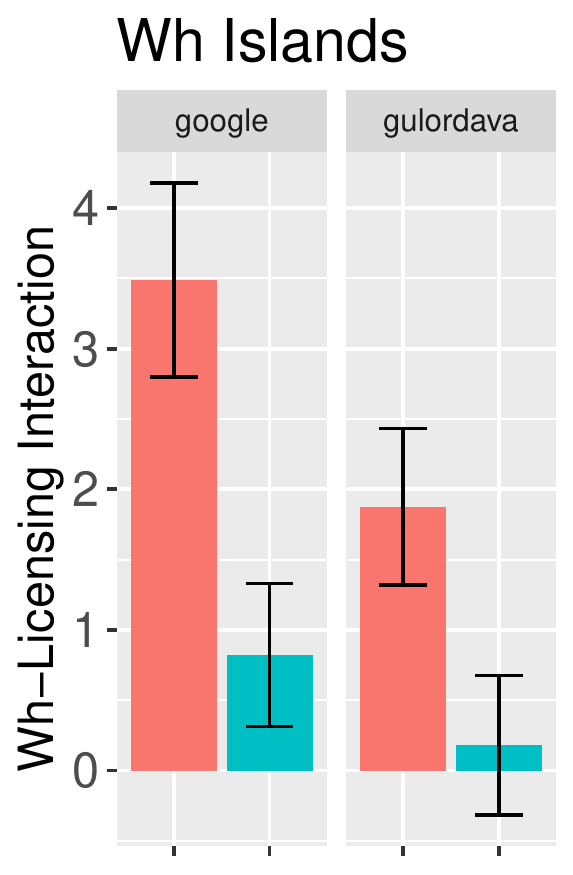}
\end{minipage}
\begin{minipage}{0.16\textwidth}
\includegraphics[width=\textwidth]{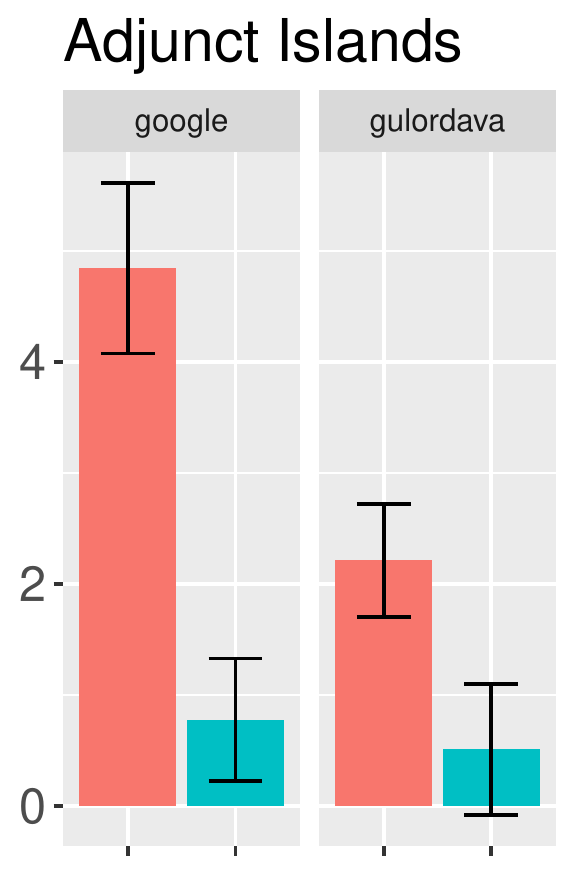}
\end{minipage}
\begin{minipage}{0.16\textwidth}
\includegraphics[width=\textwidth]{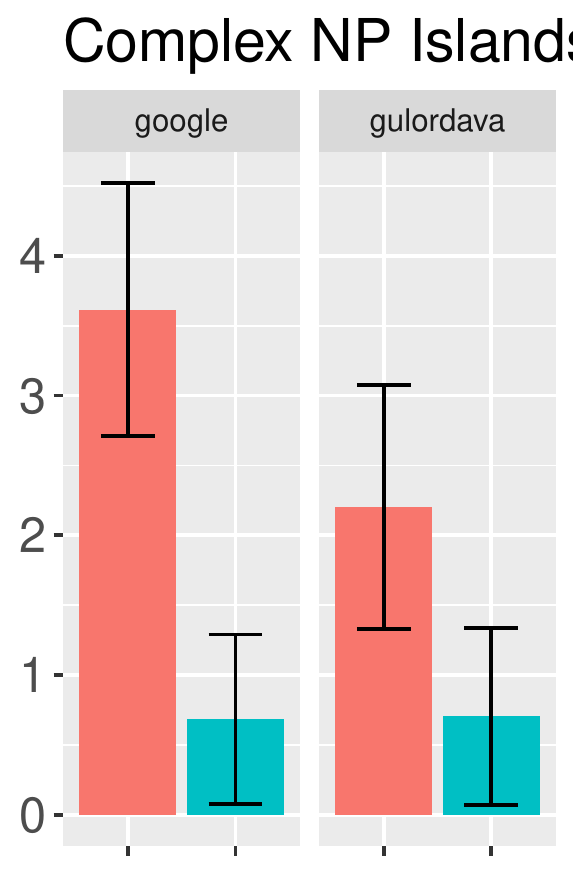}
\end{minipage}
\begin{minipage}{0.16\textwidth}
\includegraphics[width=\textwidth]{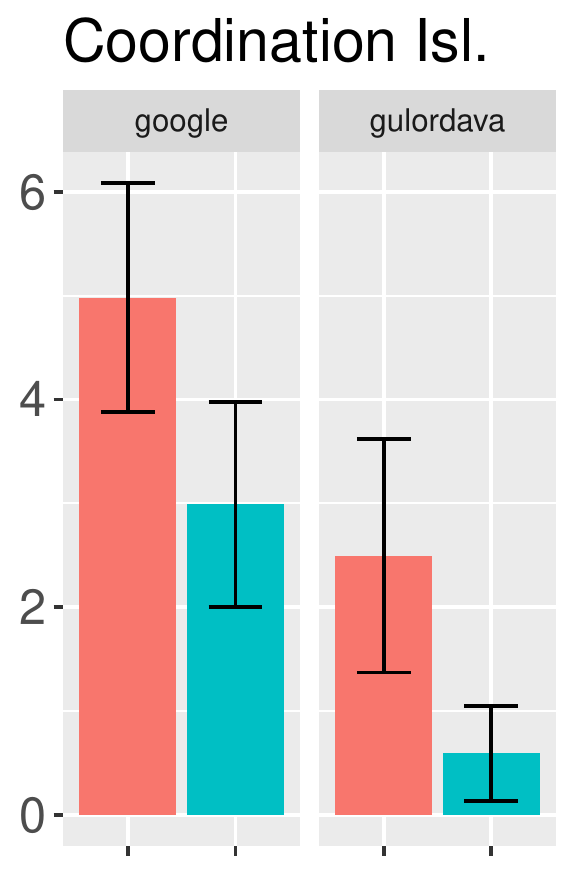}
\end{minipage}
\begin{minipage}{0.16\textwidth}
\includegraphics[width=\textwidth]{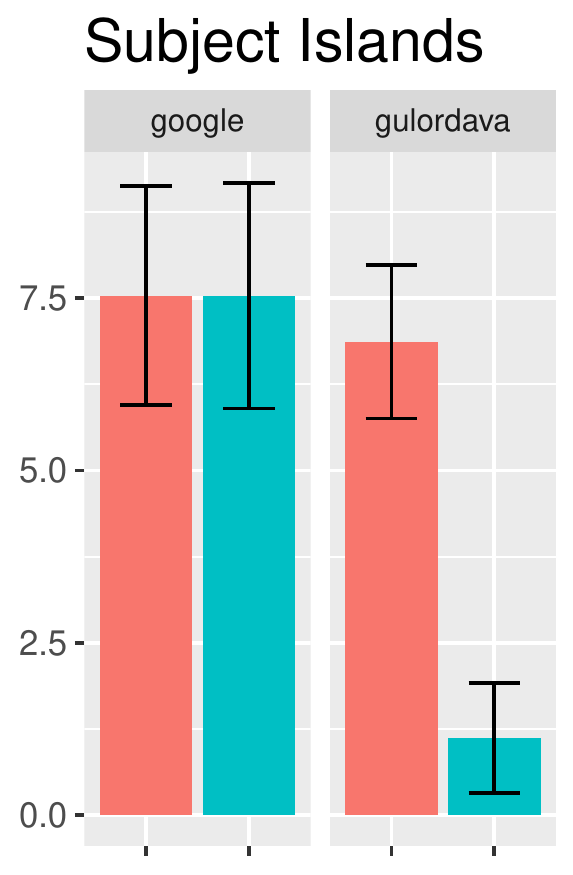}
\end{minipage}
\begin{minipage}{0.16\textwidth}
\includegraphics[width=\textwidth]{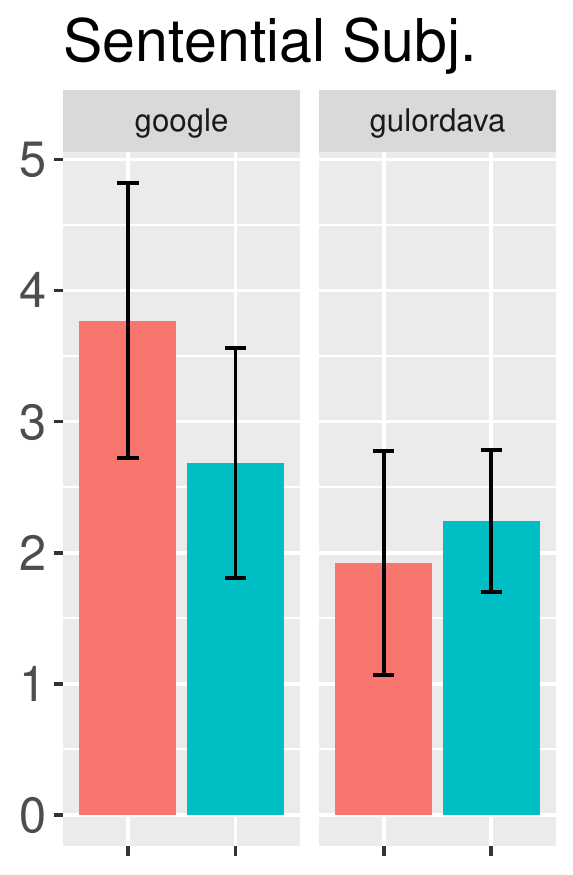}
\end{minipage}
\begin{minipage}{0.16\textwidth}
\includegraphics[width=\textwidth]{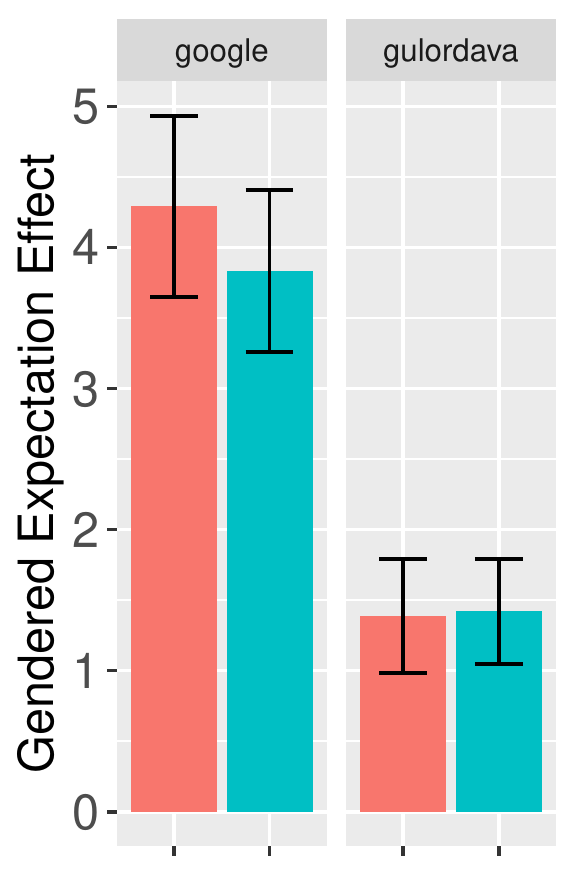}
\end{minipage}
\begin{minipage}{0.16\textwidth}
\includegraphics[width=\textwidth]{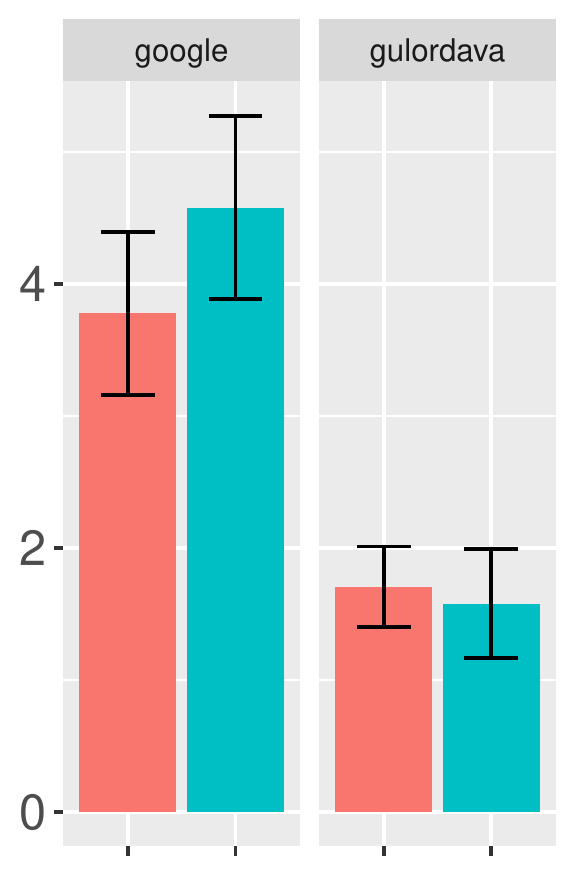}
\end{minipage}
\begin{minipage}{0.16\textwidth}
\includegraphics[width=\textwidth]{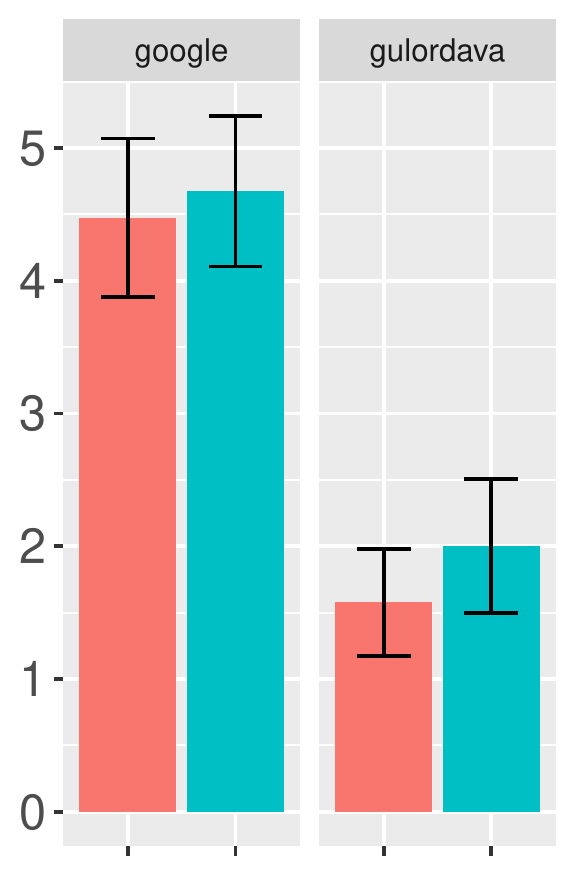}
\end{minipage}
\begin{minipage}{0.16\textwidth}
\includegraphics[width=\textwidth]{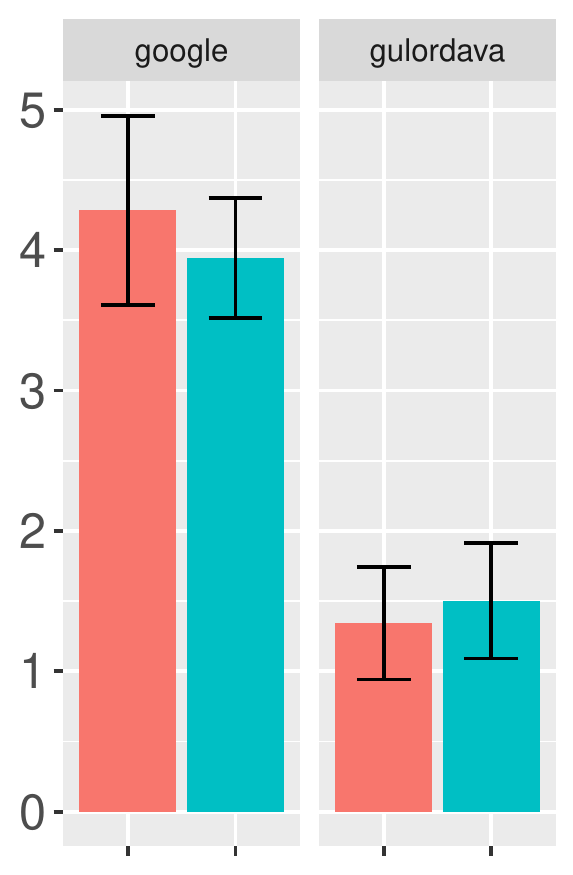}
\end{minipage}
\begin{minipage}{0.16\textwidth}
\includegraphics[width=\textwidth]{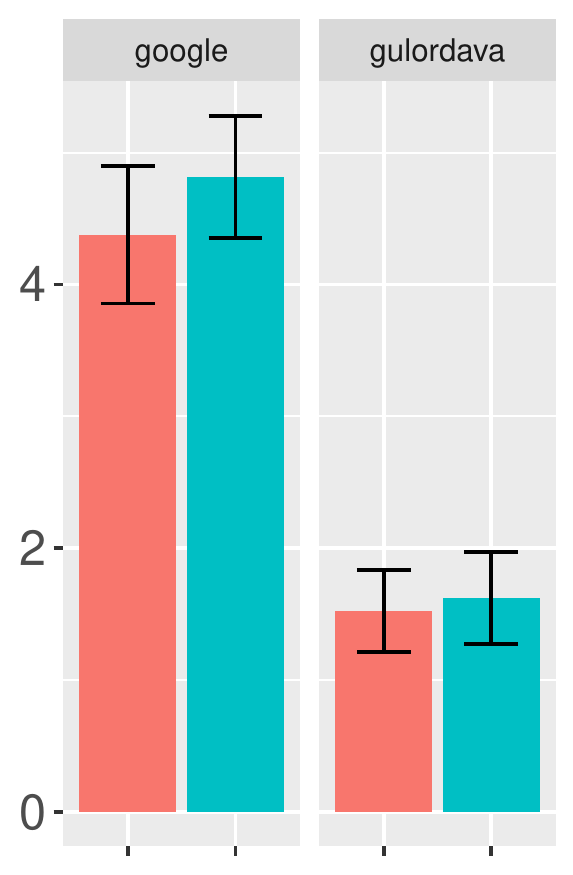}
\end{minipage}
\begin{minipage}{0.16\textwidth}
\includegraphics[width=\textwidth]{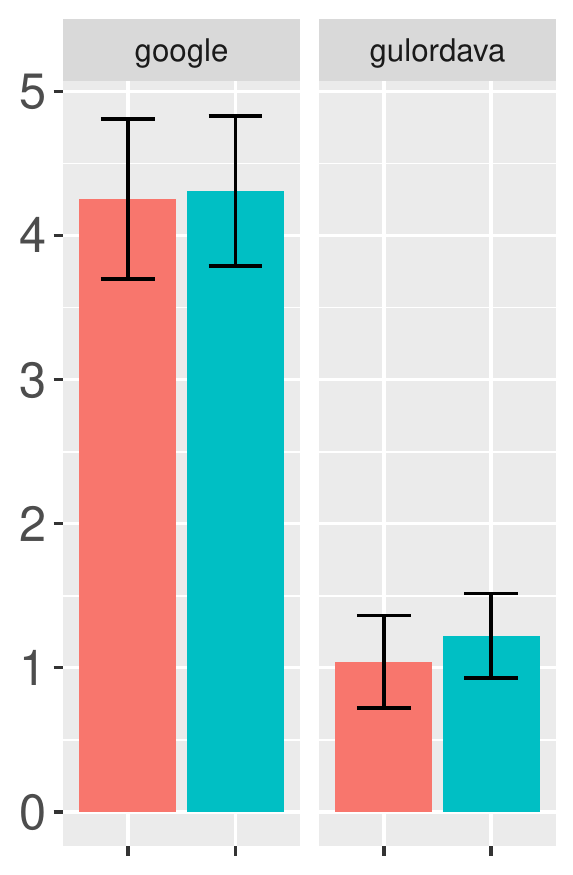}
\end{minipage}
\vspace{-0.3cm}
\caption{Effect of island construction on gender dependency.}
\label{fig:island-results}
\vspace{-0.5cm}
\end{figure*}

In order to test whether the models maintained their gender expectations through island constructions, we created six suites of experiments following the pattern of \ref{ex:gender-dep} for six of the most frequently studied islands constructions. For each of the gender expectation experiments, we created 30 variants, 15 with masculine subjects and 15 with feminine subjects and measured the surprisal at the possessive pronoun. The results are presented on the bottom row in Figure \ref{fig:island-results} alongside model performance on the filler--gap dependency for the same syntactic constructions (top row). For the filler--gap dependency, results for four islands had already been tested in \citeA{wilcox2018rnn}, which we present alongside novel results for \textit{Coordination Islands}, \textit{Sentential Subject Islands} and \textit{Left-Branch Islands}, the latter separately without a gender expectation control. For these experiments, we created between 20-24 experimental items and measured the wh-licensing interaction in the post-gap material. We take a reduction in wh-licensing interaction in island constructions and no such reduction in the gender expectation as evidence that the model has both learned the island constraint, and has applied that constraint uniquely to the filler--gap dependency.

\textbf{Wh-Islands} The wh-constraint states that the filler--gap dependency is blocked by S-nodes introduced by a wh-complimentizer, as demonstrated in the unacceptability of \ref{ex:isl-wh-bad} compared to \ref{ex:isl-wh-good}. We created experimental items following the examples in \ref{ex:isl-wh} and measured their gender expectation and filler--gap dependency (filler--gap dependency materials were taken from \citeauthor{wilcox2018rnn}).

\ex. \label{ex:isl-wh}
\a. I know who Alex said your friend insulted \_\_ \textbf{yesterday}.  \textsc{\small[control, filler--gap]} \label{ex:isl-wh-good}
\b. *I know who Alex said whether your friend insulted \_\_ \textbf{yesterday}. \textsc{\small[island, filler--gap]} \label{ex:isl-wh-bad}
\c. The actress said they insulted \{\textbf{his/her}\} friends. \textsc{\small[control, gender exp.]}
\d. The actress said whether they insulted \{\textbf{his/her}\} friends. \textsc{\small[island, gender exp.]}

The results for this experiment can be seen in the far left panel of Figure \ref{fig:island-results}, with island structures graphed in blue and non-island controls in red. We find a significant difference in licensing interaction between the island and non-island conditions for both the google and gulordava models ($p<0.001$ for both models), but no such difference in gender expectation.

\textbf{Adjunct Islands} Gaps cannot be licensed inside an adjunct clause, as demonstrated by the relative unacceptability of \ref{ex:isl-adj-good} over \ref{ex:isl-adj-bad}.

\ex. \label{ex:isl-adj}
\a. I know what the librarian placed \_\_ \textbf{on the wrong shelf}.  \textsc{\small[control, filler--gap]} \label{ex:isl-adj-good}
\b. *what the patrong got mad after the librarian placed \_\_ \textbf{on the wrong shelf}. \textsc{\small[island, filler--gap]} \label{ex:isl-adj-bad}
\c. The actress thinks they insulted \{\textbf{his/her}\} performance \textsc{\small[control, gender exp.]}
\d. The actress got mad after they insulted \{\textbf{his/her}\} performance. \textsc{\small[island, gender exp.]}

The results for this experiment can be seen in Figure \ref{fig:island-results}, second panel from the left. We find a significant reduction of wh-licensing interaction between the control and island conditions in the case of the filler--gap dependency for both models ($p<0.001$ google; $p<0.01$ gulordava; materials taken from ]\citeauthor{wilcox2018rnn}). However, we find no effect of syntactic structure on the gender effect.

\textbf{Complex NP Islands} Gaps are not licensed inside S-nodes that are dominated by a lexical head noun, as demonstrated by the relative badness of \ref{ex:isl-cnpc-bad} compareid to \ref{ex:isl-cnpc-good}.

\ex. \label{ex:isl-cnpc}
\a. I know what the actress bought \_\_ \textbf{yesterday}.  \textsc{\small[control, filler--gap]} \label{ex:isl-cnpc-good}
\b. *I know what the actress bought the painting that depicted \_\_ \textbf{yesterday}. \textsc{\small[island, filler--gap]} \label{ex:isl-cnpc-bad}
\c. The actress said they saw her \{\textbf{his/her}\} performance. \textsc{\small[control, gender exp.]}
\d. The actress said they saw the exhibit that featured \{\textbf{his/her}\} performance. \textsc{\small[island, gender exp.]}

We created items follwing the examples in \ref{ex:isl-cnpc}, with filler--gap items adopted from \cite{wilcox2018rnn}. The results from this experiment can be found in the middle-left panel of Figure \ref{fig:island-results}. We found an effect of syntactic location on wh-licensing interaction for both models ($p<0.001$ google; $p<0.01$ gulordava) but no such interaction for gender expectations.

\textbf{Coordination Islands} The coordination constraint states that a gap cannot occur in one half of a coordinate structure as demonstrated by the difference between \ref{ex:isl-coord-bad} and \ref{ex:isl-coord-good}, in which a whole conjunct has been gapped.

\ex. \label{ex:isl-coord}
\a. I know what the man bought \_\_ \textbf{at the antique shop}.  \textsc{\small[control, filler--gap]} \label{ex:isl-coord-good}
\b. *I know what the man bought the painting and \_\_ \textbf{at the antique shop}. \textsc{\small[island, filler--gap]} \label{ex:isl-coord-bad}
\c. The fireman knows they talked about \{\textbf{his/her}\} performance. \textsc{\small[control, gender exp.]}
\d.The fireman knows they talked about the football game and \{\textbf{his/her}\} performance. \textsc{\small[island, gender exp.]}

We created experimental items following the examples in \ref{ex:isl-coord}. Results can be seen in \ref{fig:island-results} center-right panel. For the filler--gap dependency, in both models there is a significant difference between the \textit{control} condition and \textit{island} conditions ($p<0.05$ for both models). These results indicate that the models have somewhat attenuated expectations for gaps when they occur in the second half of a coordinate structure. However, note that, at least for the google model, the wh-licensing interaction is significantly greater than zero, indicating that this model still maintains \textit{some} expectation for gaps in this syntactic location. For both models there is no difference in gender expectation between the \textit{control} and \textit{island} conditions).

\textbf{Subject Islands} Gaps are generally licensed in prepositional phrases, except when they occur attached to sentential subjects. We created experimental items following the examples in \ref{ex:isl-subj}, with filler--gap materials adapted from \citeauthor{wilcox2018rnn}.

\ex. \label{ex:isl-subj}
\a. I know what  \_\_ \textbf{fetched} a high price.  \textsc{\small[control, filler-gap]}
\b. *I know who the painting that depicted \_\_ \textbf{fetched} a high price. \textsc{\small[island, filler--gap]}
\c. The actress said they sold the painting by \{\textbf{his/her}\} friend. \textsc{\small[control, gender exp.]}
\d. The actress said the painting by \{\textbf{his/her}\} friend sold for a lot of money. \textsc{\small[island, gender exp.]}

The results from this experiment can be seen in Figure \ref{fig:island-results}, second panel from the right. For the filler--gap dependency, we found a significant difference between the \textit{control} and \textit{island} condition in the case of the gulordava model ($p<0.01$), but no such reduction in the case of the google model. For gender expectation, we found no significant difference between the two conditions.

\textbf{Sentential Subject Islands} The sentential subject constraint states that gaps are not licensed within an S-node that plays the role of a sentential subject. To assess whether the RNN models had learned this constraint we created items following the variants in \ref{ex:isl-sentsubj}.

\ex. \label{ex:isl-sentsubj}
\a. I know who the seniors defeated \_\_ \textbf{last week}.  \textsc{\small[control, filler--gap]}
\b. I know who for the seniors to defeat  \_\_ \textbf{will be trivial}. \textsc{\small[island, filler--gap]}
\c. The fireman knows they will save \{\textbf{his/her}\} friend. \textsc{\small[control, gender exp.]}
\d. The fireman knows for them to save \{\textbf{his/her}\} friend will be difficult. \textsc{\small[island, gender exp.]}

The results for this experiment can be seen in Figure \ref{fig:island-results}, in the far right panel. We found no decrease in gender expectation between the \textit{control} and \textit{island} conditions for either model. Likewise, for the filler--gap dependency we found no significant decrease in wh-licesning interaction between the island and non island conditions in either model. These results indicate that neither model suspends its expectations for gaps within sentential subjects.

\begin{figure}
\begin{minipage}{0.45\textwidth}
\includegraphics[width=\textwidth]{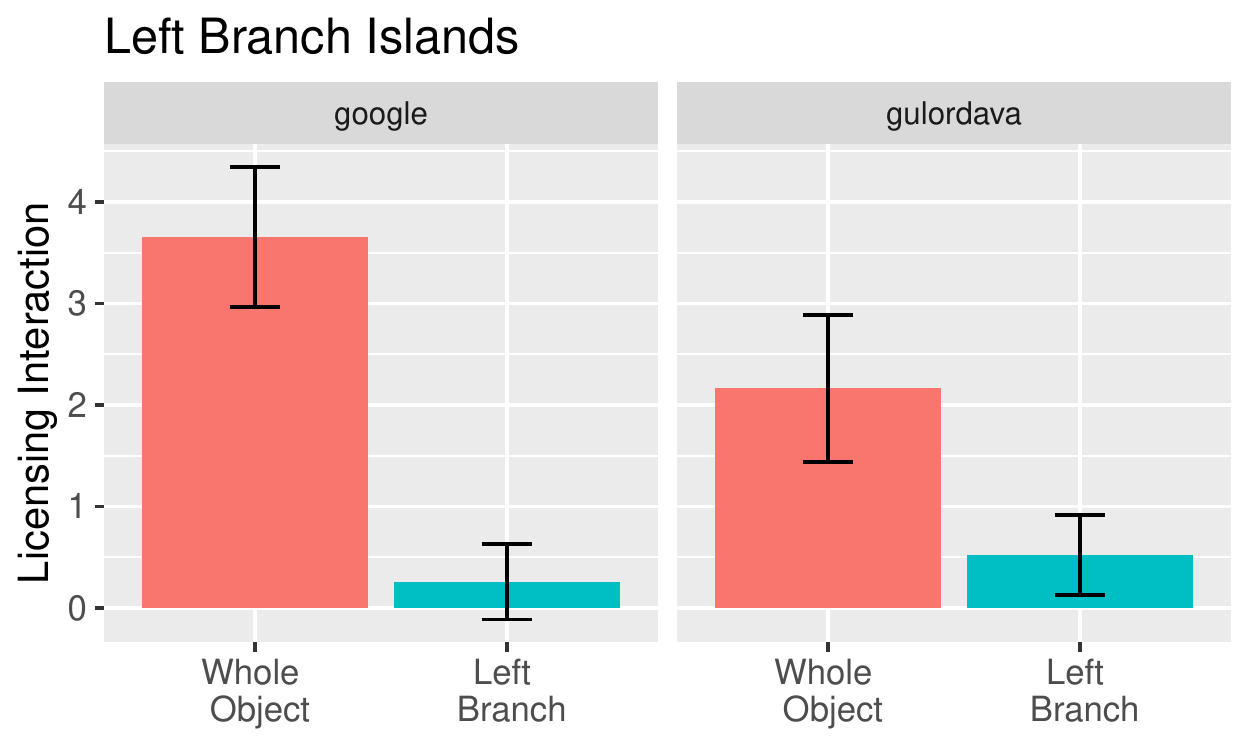}
\end{minipage}
\vspace{-0.4cm}
\caption{Left Branch Islands.}
\label{fig:left-branch}
\vspace{-0.6cm}
\end{figure}

\textbf{Left Branch Islands} The left-branch constraint states that modifiers which appear on the left branch under an NP cannot be gapped, which accounts for the relative ungrammaticality of \ref{ex:isl-leftbranch-bad} compared to \ref{ex:isl-leftbranch-good}. Because possessive pronouns cannot grammatically occur in left-branches under an NP, this experiment examines only the filler--gap dependency.  We created 20 items following the examples in \ref{ex:left-branch} and measured the wh-licensing interaction in the post-gap material.

\ex. \label{ex:left-branch}
\a.I know what color car you bought \_\_ \textbf{last week}. \textsc{\small[whole object]}  \label{ex:isl-leftbranch-good}
\b. I know what color you bought \_\_ \textbf{car last week}. \textsc{\small[left branch]} \label{ex:isl-leftbranch-bad}

The results from this experiment can be seen in Figure \ref{fig:left-branch} with experimental conditions on the x-axis and wh-licensing interaction on the y-axis. We see strong wh-licensing interaction in the two \textit{whole object} conditions, but a significant reduction in licensing interaction when the gap consists of the Adjective Phrase modifier ($p<0.001$ for the google model; $p<0.05$ for the gulordava model). This results indicate that the models have learned the left branch islands, insofar as they do not expect left-branching modifiers to be extracted without the NP to which they are attached.

For every condition tested we found that the expectation set up by gendered subjects for possessive pronouns is not affected by the pronoun's location inside island constructions. For the three novel structures, we found that the two models tested are sensitive to left branch islands and gradiently to coordination islands, but not to sentential subject islands.

\vspace{-0.2cm}
\section{Discussion}
\vspace{-0.2cm}

The filler--gap dependency has been the focus of intense research for over fifty years because it is both far reaching and tightly constrained. It can be threaded through a potentially unbounded number of sentential embeddings; yet the filler must syntactically dominate the gap and the dependency is subject to a number of highly-specific blocking `island' conditions. In this work we have shown that RNNs trained on a language modeling objective have learned both the power and the constraints imposed on this dependency. First, we provided evidence that they are able to thread the dependency through an unbounded number of sentential embeddings, and have also learned the constraints that govern the syntactic hierarchy of the filler relative to the gap. 

Second, using gender expectation effects, we have demonstrated that the models are able to thread some contextually-dependent expectations into island constructions, providing evidence that previously-observed island effects have been learned for the filler--gap dependency \emph{in particular}, and are not due to the model's inability to thread \emph{any} information into syntactic islands. In addition, we have increased the experimental coverage of island effects, demonstrating that the models were able to learn left-branch islands and gradiently learn coordination islands, but failed to learn sentential subject islands. This brings the total number of islands learned to 5/7 for the google model and 6/7 for the gulordava model. Although some of the model behavior remains strikingly unlike human acceptability judgements (in e.g. coordination islands), these experiments demonstrate that sequence models trained on a language modeling objective are able to separate natural language dependencies from each other and learn different fine-grained syntactic rules for each.

\bibliographystyle{apacite}

\bibliography{everything}

\end{document}